\documentclass{article}

\usepackage{iclr2027_conference,times}

\newif\ificlrpreprint
\iclrpreprinttrue
\ificlrpreprint
  \iclrfinalcopy
\fi

\usepackage[utf8]{inputenc}
\usepackage[T1]{fontenc}
\usepackage{hyperref}
\usepackage{url}
\usepackage{fontawesome5}
\usepackage{booktabs}
\usepackage{tabularx}
\usepackage{graphicx}
\usepackage{wrapfig}
\usepackage{flafter}
\usepackage{float}
\usepackage{amssymb}
\usepackage{nicefrac}
\usepackage{microtype}
\usepackage{xcolor}
\usepackage{tikz}
\usepackage[colorinlistoftodos,textsize=tiny]{todonotes}
\usetikzlibrary{arrows.meta,positioning}

\usepackage{colortbl}

\colorlet{benchgreen}{green!22}
\colorlet{benchred}{red!25}
\colorlet{benchmixed}{benchgreen!50!benchred}

\newcommand{\benchyes}{\cellcolor{benchgreen}$\checkmark$}
\newcommand{\benchno}{\cellcolor{benchred}$\times$}
\newcommand{\benchpartial}{\cellcolor{benchmixed}$\sim$}

\newcommand{\benchname}{\textsc{LongHarness}}
\newcommand{\Notebook}{Program Execution Tracing}
\newcommand{\Memos}{Outlier Memo Detection}

\title{\textsc{LongHarness Bench:} Stress-Testing Language Model Harnesses for Long-Context Reasoning}

\ificlrpreprint
\author{
Quang Hieu Pham\thanks{Equal contribution. $^\dagger$ Equal advising.}\hspace{0.6em}$^{\mathcal{A}}$ \enspace
Thuy Duong Nguyen\footnotemark[1]\hspace{0.6em}$^{\mathcal{A}}$ \enspace
\textbf{Jocelyn Qiaochu Chen}\footnotemark[2]\hspace{0.6em}$^{\mathcal{A}\mathcal{M}}$ \enspace
\textbf{Xi Ye}\footnotemark[2]\hspace{0.6em}$^{\mathcal{A}\mathcal{M}}$
\\[6pt]
$^{\mathcal{A}}$University of Alberta \quad
$^{\mathcal{M}}$Alberta Machine Intelligence Institute (Amii)
\\[2pt]
\texttt{\{quanghieupham, thuyduongnguyen, jocelyn.chen, xi.ye\}@ualberta.ca}
\\[2pt]
\textcolor{blue!65!black}{\faGlobe\enspace\url{https://stringnlplab.github.io/longharness/}}
}
\else
\author{Anonymous Authors}
\fi

\begin{document}

\maketitle

\ificlrpreprint
  \lhead{\sc LongHarness Bench} 
\fi

\begin{abstract}


Language-model (LM) harnesses enable LMs to operate effectively over long contexts using additional compute. However, existing long-context evaluations are insufficient for distinguishing modern harnesses, reflected by saturated accuracy across harnesses and largely similar evaluation costs. In this paper, we introduce a benchmark for evaluating both the effectiveness and efficiency of long-context harnesses. Our tasks require diverse retrieval strategies, including lexical search and semantic matching, together with strategic and adaptive reasoning over global and local context. Much of the context is semantically relevant but only a small subset is useful at each step, creating both a challenging search problem and different accuracy--cost tradeoffs across processing strategies.
For example, one task requires identifying every person satisfying several conditions using evidence scattered across documents; strategically checking the most selective condition first can narrow the search before verifying the remaining conditions. We evaluate multiple families of frontier language models with four state-of-the-art harnesses. Our benchmarks remain challenging even for strong model--harness combinations: the best reaches 68\% macro-average accuracy across four evaluation suites. More importantly, we find that the same underlying model can exhibit markedly different efficiency under different harnesses. Our results establish efficiency as an important axis for long-context evaluation and provide a testbed for developing harnesses that process context strategically rather than exhaustively.

\end{abstract}

\section{Introduction}

\begin{wrapfigure}[19]{r}{0.333\textwidth}
  \vspace{-3em}
  \centering
  \includegraphics[width=\linewidth]{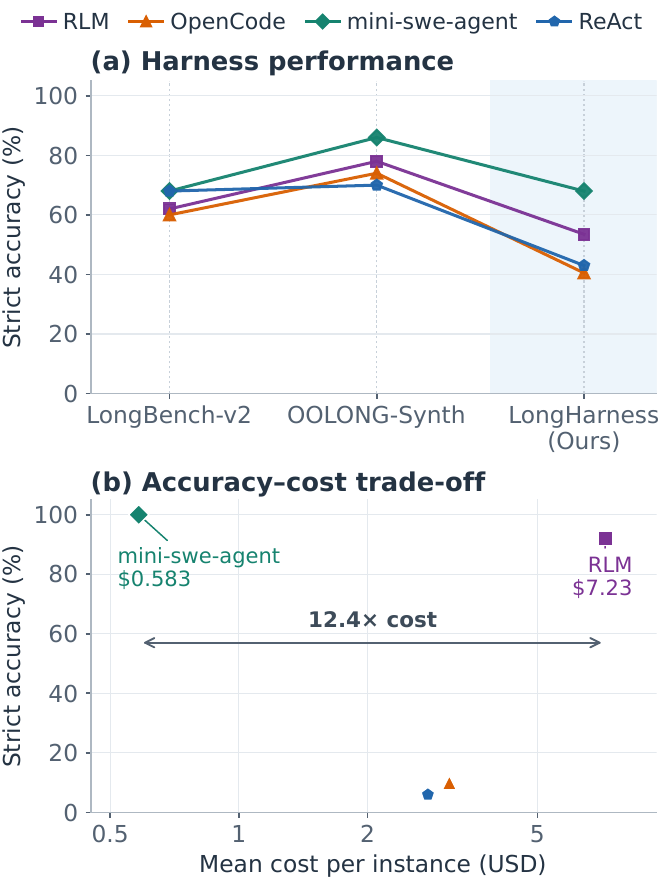}
  \vspace{-1.0em}
  \caption{\textbf{\benchname{} (ours) reveals harness differences} in both accuracy and cost.}
  \label{fig:teaser}
  \vspace{-1.2em}
\end{wrapfigure}




Language-model (LM) harnesses extend a model's ability to reason over long contexts through search, context slicing, tool use, and additional model calls~\citep{yang2024sweagent,zhang2025rlm}.  These techniques can substantially improve performance on long-context tasks while spending additional computation.  The rapid progress in harness development expose two gaps in existing long-context evaluations. First, tasks that can be solved through simple retrieval or by processing context segments independently reveal little about a harness’s ability to adapt its strategy as reasoning unfolds. The increasingly saturated accuracy of strong model–harness combinations on these benchmarks further limits their ability to distinguish harness capabilities~\citep{cao2026codingagents}. Second, accuracy alone does not capture the computational cost of harness performance: systems built around the same model may achieve similar accuracy while consuming substantially different amounts of computation (Figure~\ref{fig:teaser}; bottom). Evaluating long-context harnesses therefore requires tasks that challenge adaptive reasoning, together with measures of the computation needed to solve them.

\begin{figure}[!t]
  \centering
\includegraphics[width=1.0\linewidth,trim=0 525 10 63,clip]{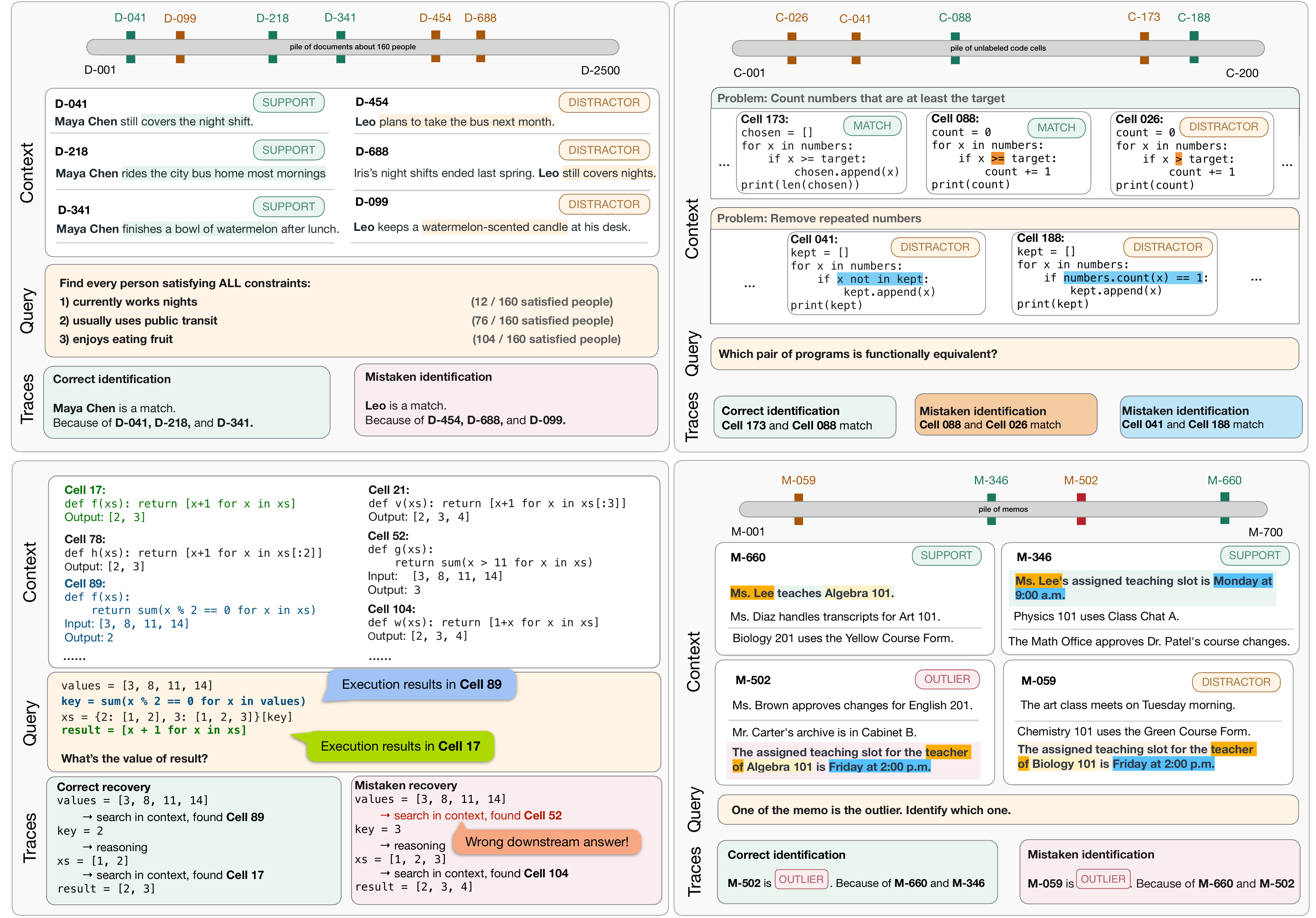}
\vspace{-2em}
  \caption{Examples from our benchmark. Top left: Constraint Solving Search combines evidence from documents. Top right: Equivalent
  Program Pair Search distinguishes equivalent programs from near matches. Bottom left: \Notebook{} requires
  reading prior outputs and matching code semantics. Bottom right: \Memos{}
  identifies a conflict using supporting memos. 
  }
  \vspace{-1.5em}
  \label{fig:tasks}
\end{figure}

We introduce \benchname{}, a benchmark comprising four diverse long-context tasks that require challenging retrieval and adaptive reasoning while admitting multiple solution strategies with different computational costs. Specifically, retrieval requires locating useful evidence among many semantically relevant candidates, where lexical overlap or semantic similarity alone cannot reliably identify the correct evidence. Reasoning requires combining information across the context over multiple steps, with intermediate findings often determining what to inspect next. These demands also create opportunities for efficiency: harnesses can strategically narrow their searches, prioritize informative evidence, and reuse intermediate results to avoid unnecessary processing. %


For example, our Constraint Solving Search task (Figure~\ref{fig:tasks}, top left) asks agents to identify every person satisfying several conditions using evidence scattered across a large collection of documents. The task calls for both lexical and semantic search: entity names help locate passages about a candidate, while conditions such as ``enjoying fruit'' require recognizing evidence expressed in different terms, such as a ``fondness for watermelon''. Since many people satisfy most but not all conditions, agents must combine evidence across documents and check each candidate against the full query. An exhaustive strategy could check every condition for every person, whereas a more selective strategy could first identify the condition satisfied by the fewest people, then verify the remaining conditions only for those candidates. Both approaches can reach the correct answer, but they require different amounts of processing, allowing the task to distinguish accuracy and efficiency across model--harness combinations.


We evaluate direct inference and four agentic harnesses---OpenCode,
mini-swe-agent, RLM, and ReAct
\citep{opencode2026,yang2024sweagent,zhang2025rlm,yao2023react}---on all four
\benchname{} tasks. Our evaluation covers five models: GPT-5.6-sol, Gemini 3.8
Flash, GLM-5.3, Qwen3.8-27B, and Kimi-K2.6
\citep{openai2026gpt56sol,glm5team2026glm5,qwen2026qwen38,moonshot2026kimik26}.
For comparison, we also evaluate GPT-5.6-sol configurations on OOLONG-Synth and
LongBench-v2 \citep{bertsch2025oolong,bai2024longbenchv2}. Figure~\ref{fig:teaser}
shows that, with GPT-5.6-sol as a backbone, harness accuracy varies more widely on
\benchname{} than on LongBench-v2 or OOLONG-Synth
(Figure~\ref{fig:teaser}a). On our Outlier Memo Detection task, RLM achieves accuracy
comparable to mini-swe-agent but costs $12.4\times$ as much per instance, revealing
an efficiency difference hidden by accuracy alone (Figure~\ref{fig:teaser}b).
The best configuration
on our suite reaches 68\% macro-average accuracy, and no harness leads across all
four tasks. For GPT-5.6-sol, every harness helps on some tasks and hurts on others,
while additional inference can sharply increase cost without improving accuracy.
Compared with the two existing benchmarks, our suite more clearly reveals how
harness choice affects both accuracy and computational cost. Further analysis
reveals harness failures to reuse evidence, verify candidates, and reach an answer
within budget. These findings highlight substantial room to improve the
reliability and efficiency of long-context reasoning, and establish
\benchname{} as a testbed for this research.

\section{Background and Motivation}
\label{sec:background}

\begin{table}[t]
      \centering
      \footnotesize
      \setlength{\tabcolsep}{4pt}
      \renewcommand{\arraystretch}{1.2}
      \begin{tabularx}{\linewidth}{@{}Xccccc@{}}
          \toprule
          & \shortstack{LongBenchV2}
          & HELMET
          & LongProc
          & \shortstack{Oolong}
          & \textbf{\benchname{}} \\
          \midrule
          Diverse, adaptive retrieval demand
          & \benchpartial & \benchpartial & \benchno
          & \benchno & \benchyes \\

          Semantically confusable evidence
          & \benchno & \benchyes & \benchpartial
          & \benchpartial & \benchyes \\

         Extensive reasoning steps
  & \benchno & \benchno & \benchyes
  & \benchyes & \benchyes \\

    Step-dependent retrieval & \benchno & \benchno & \benchyes
  & \benchno & \benchyes \\
          Multiple strategies with different costs
          & \benchno & \benchpartial & \benchno
          & \benchpartial & \benchyes \\

          \midrule
          Wide range of SoTA system accuracy
          & \benchno & \benchno & \benchno
          & \benchno & \benchyes \\

          Wide range of SoTA system cost
          & \benchno & \benchno & \benchno
          & \benchpartial & \benchyes \\
          \bottomrule
      \end{tabularx}
      \caption{Comparison of long-context benchmarks along properties relevant
      to evaluating LM harnesses.
      $\checkmark$: supported; $\sim$: partially supported;
      $\times$: not supported.
      Diverse, adaptive retrieval demand means choosing an appropriate way to
      access context, such as lexical search, semantic retrieval, or direct
      reading, as information needs change during reasoning. Step-dependent retrieval means that
      intermediate findings determine what must be retrieved next. The final
      two rows indicate whether model--harness configurations are widely
      separated in accuracy and cost.}
      \label{tab:benchmark-comparison}
  \end{table}
  
We describe how language-model harnesses support long-context reasoning and discuss what existing benchmarks reveal about their capabilities. We then identify requirements for a more comprehensive evaluation of harness accuracy and efficiency, motivating the design of \benchname{}.

\subsection{Language-Model Harnesses for Long-Context Reasoning}

  We use \emph{LM harness} to mean the external control layer
  around a base model that manages its context, tools, and sequence of
  interactions~\citep{huang2026memoharness}.
  For long-context reasoning, a harness allows the model to process a large
  input through successive interactions, using the results of earlier steps to guide later ones. Coding-agent 
  harnesses such as OpenCode and mini-swe-agent let models work with context stored in files, while
  RLM exposes the context as a programmatically accessible object that can be processed through recursive
  model calls~\citep{opencode2026,yang2024sweagent,zhang2025rlm,cao2026codingagents}. These mechanisms can
  improve accuracy through additional model calls and repeated inspection of the context, but this extra
  processing also increases computational cost. Assessing their value therefore requires measuring
  both the accuracy they achieve and the computation they consume.

  \paragraph{Harnesses enable flexible retrieval and reasoning.}
  A harness lets a model choose how to access context according to its current information
  needs. Lexical search can locate known identifiers, while semantic retrieval can find
  relevant passages expressed in different terms. As reasoning progresses, the model can
  use intermediate findings to refine its searches and retain useful evidence for later
  subproblems. These choices allow different strategies for the same task: a model may
  inspect the context exhaustively or use a sequence of targeted searches to reach the
  answer with less computation.

  \subsection{Existing Long-Context Benchmarks}
  \label{sec:existing_dataset}
  We briefly review popular long-context benchmarks and discuss their limitations in evaluating LM harnesses. LongBench v2 and HELMET cover broad long-context applications \citep{bai2024longbenchv2,yen2025helmet}. Their recall-oriented tasks emphasize finding a small number of relevant pieces of information and combining them through a few reasoning steps. Such tasks test evidence recovery, but place less emphasis on extensive reasoning in which intermediate findings repeatedly change what must be retrieved next. They therefore provide a limited test of a harness's ability to adapt its retrieval strategy over a longer reasoning process.

  LongProc and Oolong emphasize extended processing. LongProc \citep{ye2025longproc} tests long procedural
  generation, but provides detailed procedures for models to execute. It emphasizes faithful execution, leaving strategy choice
  less directly tested. Oolong-Synth \citep{bertsch2025oolong} requires aggregation across many
  pieces of information, but admits a broadly applicable divide-and-conquer strategy: classify individual records independently, then
  aggregate the results. This strategy processes context without repeatedly adapting retrieval to intermediate conclusions. Thus,
  both benchmarks impose substantial processing demands but provide limited coverage of tasks where difficult retrieval and
  strategic choices jointly determine accuracy and cost.

  These gaps motivate tasks combining diverse, adaptive retrieval with extensive reasoning and multiple viable solution
  strategies. Such tasks test whether a harness can choose an effective approach that reduces computational cost while
  maintaining accuracy. Table~\ref{tab:benchmark-comparison} summarizes these properties.

\section{Tasks in \benchname{}}
\label{sec:dataset}
We first outline our general design desiderata, then introduce the four tasks and their instructions. Appendix~\ref{app:construction} provides the detailed construction procedures.

\paragraph{Design Desiderata.}
Our design follows the five task properties in Table~\ref{tab:benchmark-comparison}.
To challenge both retrieval and reasoning, tasks should require diverse, adaptive
retrieval, with agents choosing among lexical search, semantic retrieval, and
direct reading as their information needs change. Contexts should contain
semantically confusable evidence that requires verification, and solutions should
involve extensive reasoning with intermediate findings guiding subsequent
retrieval.

To expose differences in efficiency, tasks should also admit multiple solution
strategies with different computational costs. Selective search and reuse of
intermediate results should offer cheaper alternatives to exhaustive processing.
Together, these requirements aim to reveal differences in accuracy and computational cost across model–harness combinations, corresponding to the final two rows of 
Table~\ref{tab:benchmark-comparison}. Each task description below explains how its
construction meets these requirements.


\subsection{Constraint Solving Search.}
Given documents describing a group of people, the agent must find every person
satisfying all conditions in a query. Each instance contains roughly 120K tokens
of varied documents about 160 people and a query with three to five conditions,
such as habits, current circumstances, or relationships with others. Exactly
five people qualify, but their evidence is scattered across documents and mixed
with close counterexamples. In Figure~\ref{fig:tasks} (top left), Maya's
night work, transit habit, and fruit eating appear in separate passages, whereas
Leo's planned bus trip and watermelon-scented candle do not establish the
corresponding habits. The answer must contain exactly the five qualifying IDs.

An efficient strategy first identifies the condition satisfied by the fewest
people, then checks the remaining conditions only for those candidates. We
deliberately include one selective condition and one common condition without
revealing which is which, along with people who satisfy all but one condition.
A solver can instead check every condition for all 160 people, but this requires
substantially more retrieval and reasoning. The task tests whether a harness can
interpret evidence precisely and narrow its candidate set through a useful search
order before full verification.

\subsection{Equivalent Program Pair Search.}
Given a collection of programs, the agent must identify every pair that produces
the same outputs for the same valid inputs. Each instance contains roughly 100K
tokens of code across 200 Python programs adapted from APPS
\citep{hendrycks2021apps}. Program statements and visible input--output examples
are omitted, and each correct implementation is surrounded by behaviorally
different mutants that solve an apparently similar problem. Matching programs
can therefore look less alike than incorrect candidates. The required output is
the complete set of matching ID pairs; the number of pairs is not disclosed.
Our answer key treats two programs as matching when both pass the tests for the
original problem, which does not prove equivalence on every possible input.

An efficient strategy first infers the purpose of each program and groups likely
matches, then uses distinguishing inputs to eliminate behaviorally different
variants within each group. In Figure~\ref{fig:tasks} (top right), Cells~173
and 088 count values greater than or equal to a target, whereas Cell~026 counts
only values strictly greater than it; an input containing the target separates
them. Grouping avoids comparing all program pairs but leaves many near-matches.
The task therefore tests whether a harness
can narrow candidates in stages and reserve detailed verification for the most
plausible pairs.

\subsection{\Notebook{}.}
\label{sec:notebook}
This task (Figure~\ref{fig:tasks}, bottom left) asks the agent to recover the result
of an eight-step query program from 320 shuffled records of previous Python
computations. Each cell records its ID, function, example call, and output.
For every query step, the agent must find a cell whose function
is semantically equivalent and whose recorded call uses the required input.
Earlier outputs determine later inputs, making retrieval step-dependent. Exact
input values can narrow the search through literal matching, but deliberately
confusable functions may agree on the displayed example while differing on
other inputs. The answer contains the eight selected cell IDs and the exact
final result in JSON.

A solver can recompute each query step and locate supporting cells, or use each
recovered output as the next search key, verify the matching function, and reuse
its result. The latter strategy
avoids repeating computations already present in the context, but requires
careful state tracking and semantic verification at every step. A wrong match
also changes later search keys and redirects the remaining trace. The task tests
whether a harness can reliably reuse prior computation.

%
%

\subsection{\Memos{}.}
\label{sec:memos}
The agent must identify memos whose claims contradict relationships established
elsewhere in the collection. Each
instance contains 700 labeled memos with 2,500 statements about people,
organizations, objects, and their relationships. Exactly 15 memos contain a
conflicting claim, but no contradiction is visible in isolation. In
Figure~\ref{fig:tasks} (bottom right), one memo says that Ms.~Lee teaches Algebra
101 and another assigns her Monday at 9 a.m.; together, they contradict a memo
placing the course teacher on Friday at 2 p.m. Paraphrases and closely related
but consistent claims make topical similarity insufficient. The answer must
contain exactly the 15 conflicting memo IDs.

An efficient strategy reads the packet once, organizes its statements into a
reusable table or graph of relationships, and combines pairs of facts to check
every derived claim. This representation lets the solver reuse the same evidence
across many checks. A less efficient strategy searches separately for evidence
related to each suspicious claim. This can reread the same memos and miss
contradictions that become apparent only after information from multiple memos
is combined. The task therefore exposes the cost of repeated local retrieval
relative to building a global structure once.

\section{Results}
\label{sec:evaluation}

Our evaluation examines whether \benchname{} challenges current
model--harness combinations and reveals
differences in their computational efficiency.

\subsection{Experimental Setup}
\label{sec:experimental-setup}

\paragraph{Models and harnesses.}
We evaluate GPT-5.6-sol, Gemini 3.8 Flash, Qwen3.8-27B, GLM-5.3, and Kimi-K2.6
using direct long-context inference and four agentic harnesses: RLM, OpenCode,
mini-swe-agent, and ReAct \citep{yao2023react}. GPT-5.6-sol,
Gemini 3.8 Flash, and GLM-5.3 use high reasoning effort; Qwen3.8-27B uses
medium effort; and Kimi-K2.6 uses its reasoning-enabled mode. Our ReAct
harness exposes custom tools for semantic search, batched nearest-neighbor
retrieval and document reading. We use
\texttt{BAAI/bge-m3} \citep{bge-m3} as the embedding
model for ReAct.

\paragraph{Evaluation protocol.}
Each instance has a 3M-token budget, counting cumulative input and output across
model calls. We report task-specific exact accuracy. \Notebook{} requires both
the exact final query result and the exact cited cells. \Memos{} requires the exact set of 15 memo IDs;
Constraint Solving Search, the exact person-ID set; and Equivalent Program Pair
Search, the exact set of five unordered pairs. Table~\ref{tab:main-results}
reports accuracy and estimated cost per instance. Appendix~\ref{app:detailed-results}
reports mean input and output tokens; Appendix~\ref{app:api-rates}
lists the API rates used to estimate cost.


\subsection{Main Results}

\paragraph{The benchmark remains challenging.}
The best configuration, GPT-5.6-sol with mini-swe-agent, reaches only 68\%
macro-average accuracy (Table~\ref{tab:main-results}). The strongest non-GPT
configuration (GLM-5.3 with mini-swe-agent) reaches 42.5\%, while Kimi-K2.6
never exceeds 1.5\%. Difficulty also varies by task: the best accuracy ranges
from 100\% on \Memos{} to 52\% on Equivalent Program Pair Search, and no
configuration leads across all four tasks.

\newcommand{\accost}[2]{#1\%~(\$#2)}

\begin{table*}[!tp]
  \caption{Exact accuracy and estimated cost across four suites. Each cell
  reports accuracy (\%) followed by USD per-instance cost in parentheses;
  macro averages the four task accuracies and per-task mean costs.
  Appendix~\ref{app:detailed-results} gives token counts and
  Appendix~\ref{app:api-rates} gives the API rates.}
  \label{tab:main-results}
  \centering
  \scriptsize
  \setlength{\tabcolsep}{3pt}
  \begin{tabularx}{\textwidth}{@{}ll*{5}{>{\centering\arraybackslash}X}@{}}
    \toprule
    Model & Harness & Constraint & Memos & Program & Pairs & Macro avg. \\
    \midrule
    GPT-5.6-sol & Direct         & \accost{58}{0.690} & \accost{44}{1.17} & \accost{94}{0.648} & \accost{44}{0.780} & \accost{60.0}{0.822} \\
                & RLM            & \accost{44}{0.659} & \accost{92}{7.23} & \accost{76}{1.47} & \accost{2}{7.86} & \accost{53.5}{4.30} \\
                & OpenCode       & \accost{56}{1.72} & \accost{10}{3.11} & \accost{96}{1.37} & \accost{0}{5.41} & \accost{40.5}{2.90} \\
                & mini-swe-agent & \accost{42}{0.320} & \accost{100}{0.583} & \accost{88}{1.02} & \accost{42}{1.34} & \accost{68.0}{0.816} \\
                & ReAct & \accost{32}{1.40} & \accost{6}{2.77} & \accost{82}{0.866} & \accost{52}{2.68} & \accost{43.0}{1.93} \\
    \midrule
    GLM-5.3 & Direct         & \accost{0}{0.728} & \accost{0}{1.09} & \accost{38}{0.785} & \accost{0}{1.19} & \accost{9.5}{0.948} \\
            & RLM            & \accost{34}{2.67} & \accost{16}{7.84} & \accost{58}{3.34} & \accost{0}{8.06} & \accost{27.0}{5.48} \\
            & OpenCode       & \accost{48}{1.94} & \accost{20}{3.52} & \accost{84}{4.22} & \accost{0}{3.64} & \accost{38.0}{3.33} \\
            & mini-swe-agent & \accost{66}{1.55} & \accost{58}{3.00} & \accost{46}{3.65} & \accost{0}{3.80} & \accost{42.5}{3.00} \\
            & ReAct          & \accost{14}{2.78} & \accost{0}{4.12} & \accost{34}{2.59} & \accost{4}{3.26} & \accost{13.0}{3.19} \\
    \midrule
    Qwen3.8-27B & Direct         & \accost{0}{0.356} & \accost{0}{0.292} & \accost{0}{0.363} & \accost{0}{0.341} & \accost{0.0}{0.338} \\
                 & RLM            & \accost{14}{2.50} & \accost{2}{7.23} & \accost{14}{5.78} & \accost{0}{7.13} & \accost{7.5}{5.66} \\
                 & OpenCode       & \accost{46}{6.50} & \accost{16}{6.32} & \accost{92}{8.59} & \accost{0}{6.83} & \accost{38.5}{7.06} \\
                 & mini-swe-agent & \accost{52}{2.60} & \accost{14}{6.54} & \accost{66}{5.30} & \accost{0}{6.84} & \accost{33.0}{5.32} \\
                 & ReAct & \accost{0}{3.37} & \accost{0}{5.93} & \accost{0}{1.79} & \accost{0}{4.39} & \accost{0.0}{3.87} \\
    \midrule
    Gemini 3.8 Flash & Direct         & \accost{12}{0.166} & \accost{0}{0.158} & \accost{28}{0.170} & \accost{0}{0.154} & \accost{10.0}{0.162} \\
                     & RLM            & \accost{20}{0.505} & \accost{22}{0.729} & \accost{0}{0.478} & \accost{0}{1.14} & \accost{10.5}{0.713} \\
                     & OpenCode       & \accost{0}{0.439} & \accost{28}{0.515} & \accost{18}{0.478} & \accost{0}{0.452} & \accost{11.5}{0.471} \\
                     & mini-swe-agent & \accost{4}{0.611} & \accost{64}{0.528} & \accost{32}{0.507} & \accost{0}{0.554} & \accost{25.0}{0.550} \\
                     & ReAct & \accost{0}{0.428} & \accost{30}{0.764} & \accost{26}{0.428} & \accost{28}{0.538} & \accost{21.0}{0.540} \\
    \midrule
    Kimi-K2.6 & Direct         & \accost{0}{0.670} & \accost{0}{0.546} & \accost{0}{0.696} & \accost{0}{0.631} & \accost{0.0}{0.636} \\
              & RLM            & \accost{2}{2.41} & \accost{0}{16.3} & \accost{0}{3.92} & \accost{0}{3.96} & \accost{0.5}{6.65} \\
              & OpenCode       & \accost{0}{3.86} & \accost{0}{5.64} & \accost{6}{8.99} & \accost{0}{4.10} & \accost{1.5}{5.65} \\
              & mini-swe-agent & \accost{0}{3.15} & \accost{0}{4.22} & \accost{2}{5.08} & \accost{0}{3.65} & \accost{0.5}{4.03} \\
              & ReAct          & \accost{0}{3.14} & \accost{0}{2.72} & \accost{0}{3.42} & \accost{0}{2.16} & \accost{0.0}{2.86} \\
    \bottomrule
  \end{tabularx}
\end{table*}


\begin{figure}[t]
  \centering
  \includegraphics[width=0.85\linewidth, trim= 30 20 10 0]{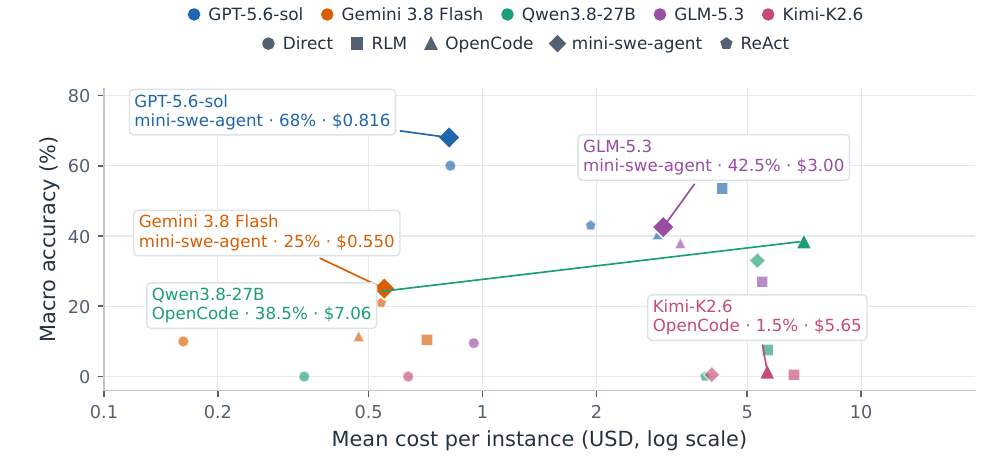}
  \caption{Macro-average accuracy and estimated cost across the four tasks
  (Table~\ref{tab:main-results}). Color denotes model; shape denotes harness.
  Callouts and larger markers mark each model's best accuracy. ReAct appears where available.}
  \label{fig:performance-cost}
\end{figure}

\begin{figure}[t]
  \centering
  \includegraphics[width=0.85\linewidth, trim= 30 20 10 0 clip ]{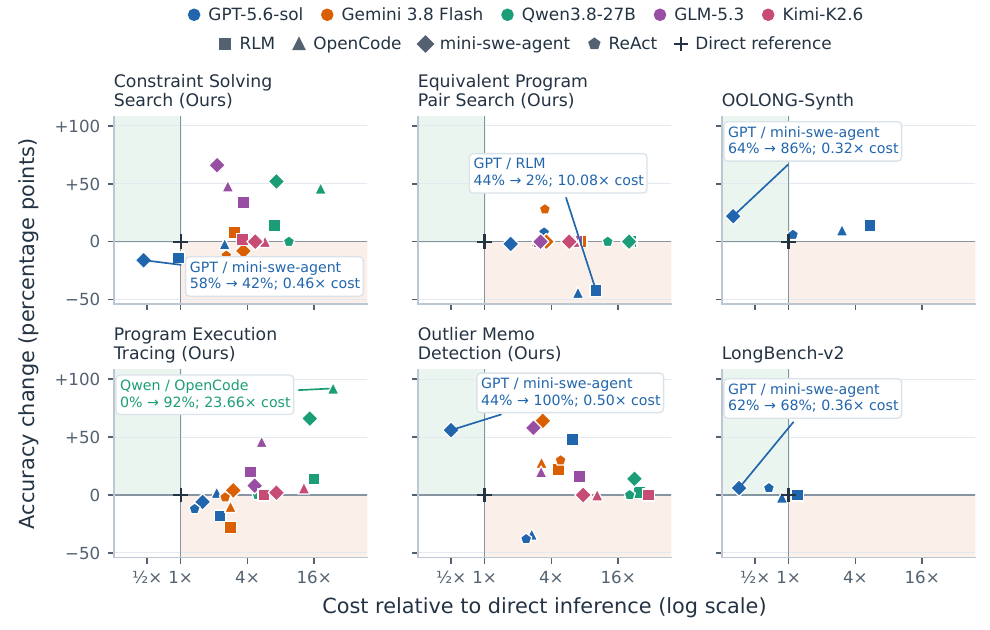}
  \caption{Harness effects relative to direct inference on OOLONG-Synth,
  LongBench-v2, and our suites.
  External benchmarks use GPT-5.6-sol. The reference is zero accuracy change
  and $1\times$ cost. Green denotes higher accuracy at lower cost;
  pink denotes lower accuracy at higher cost.}
  \label{fig:harness-task-effects}
\end{figure}

\paragraph{Similar accuracy can have different costs.}
For a fixed model, comparable accuracy can involve substantially different
resource use. On \Notebook{}, GPT-5.6-sol direct inference and OpenCode
reach 94\% and 96\% accuracy, respectively, but use 0.121M and 1.29M input
tokens and cost \$0.648 and \$1.37 per instance. On Constraint Solving
Search, direct inference reaches 58\% at \$0.690, compared with 56\% at
\$1.72 for OpenCode. Thus, small differences in accuracy can accompany much
larger differences in tokens and cost. Figure~\ref{fig:performance-cost}
compares macro accuracy and cost, while Figure~\ref{fig:harness-task-effects}
shows per-task changes relative to direct inference.

\subsection{Model--Harness Interplay}

\paragraph{Harness gains depend on the model and task.}
Harnesses can perform substantially worse than direct inference, even when
they help the same model elsewhere. For GPT-5.6-sol, OpenCode slightly
improves \Notebook{} accuracy from 94\% to 96\%, but reduces accuracy from
44\% to 10\% on \Memos{} and from 44\% to 0\% on Equivalent Program Pair
Search. Mini-swe-agent attains the highest macro accuracy (68\%, versus 60\%
for direct inference) by improving \Memos{} from 44\% to 100\%, yet it lowers
Constraint Solving Search from 58\% to 42\%. The preferred harness varies by
task; none uniformly improves on direct inference.

\paragraph{Weaker models tend to gain more from harnesses.}
Each model's best harness generally yields larger gains for models with
lower direct accuracy. Qwen3.8-27B rises from 0\% direct accuracy to
38.5\%, and GLM-5.3 rises from 9.5\% to 42.5\%, corresponding to gains of
38.5 and 33 percentage points. Gemini 3.8 Flash improves by 15 points, from
10\% to 25\%, whereas GPT-5.6-sol improves by 8 points, from 60\% to 68\%.
These gains substantially narrow the gap between model families: GLM-5.3 with
mini-swe-agent (42.5\%) and Qwen3.8-27B with OpenCode (38.5\%) are comparable
to GPT-5.6-sol with OpenCode (40.5\%) or ReAct (43.0\%), despite
their much lower direct accuracy.

\paragraph{More inference does not ensure success.}
Several configurations cost more while achieving lower accuracy. On
Equivalent Program Pair Search, GPT-5.6-sol RLM reaches 2\% at a reported
cost of \$7.86 per instance, compared with 44\% at \$0.780 for
direct inference and 52\% at \$2.68 for ReAct. On \Memos{},
GPT-5.6-sol RLM reaches 92\% at \$7.23,
while mini-swe-agent reaches 100\% at \$0.583. Additional inference is
neither necessary nor sufficient for higher accuracy.

\subsection{Harness Mechanisms and Trade-offs}
Harnesses use extra model calls to reduce the context considered at each step.
Figure~\ref{fig:harness-task-effects} shows that the benefit depends on how much
these calls simplify the remaining work.

\paragraph{File-and-shell agents: OpenCode and mini-swe-agent.}
File and shell access supports both targeted retrieval and scripted
verification, with different accuracy--cost trade-offs. On \Notebook{},
OpenCode uses each step's recorded output to search for cells that take that
value as input, then verifies which cell implements the next query step.
With GPT-5.6-sol, this raises accuracy only from 94\% to 96\% over an already
strong direct-inference baseline, while increasing cost from \$0.648 to \$1.37.
On \Memos{}, mini-swe-agent reduces repeated model reading by collecting facts
into a table and using a local script to combine them and check claims.
With the same model, it reaches 100\% accuracy at \$0.583, compared with
44\% at \$1.17 for direct inference. OpenCode often rereads the packet or uses
separate model calls to audit overlapping memos, reaching only 10\% at
\$3.11. Targeted retrieval thus offers a small accuracy gain on execution
tracing, while scripted verification improves both accuracy and cost on memo
detection.

\paragraph{Programmatic decomposition: RLM.}
RLM filters the long prompt into smaller pieces, sends them through additional
sub-model calls, and asks the main model to combine the results. On \Memos{}, this
recovers many outliers (92\%) but costs \$7.23, versus 100\% at \$0.583 for
mini-swe-agent, which checks the packet with one local program. On Equivalent
Program Pair Search, RLM uses its subcalls to infer each program's purpose and
form broad problem families. This is a useful first filter, but our construction
places many behaviorally different mutants in each family. An efficient solver
must then use distinguishing inputs to prune each family before verifying the
remaining pairs. RLM spends much of its recursion on the initial grouping and
reaches the final comparison with too many candidates, yielding 2\% at \$7.86
versus 44\% at \$0.780 for direct inference. Here decomposition reduces the
context per call without sufficiently reducing the final decision space.

\paragraph{Embedding retrieval: ReAct.}
Semantic retrieval is most useful when a retrieved item is itself a plausible
comparison candidate. This holds in Equivalent Program Pair Search: querying
with one program can surface stylistically different implementations with the
same apparent purpose, turning a search over 200 programs into a smaller
neighborhood. The model must still reject near-mutants, but this shortlist
raises GPT-5.6-sol from 44\% at \$0.780 to 52\% at \$2.68. By contrast,
our \Memos{} task cannot be solved by judging each memo in isolation. The incorrect
memo is revealed only by combining facts from two other memos. Similarity search often returns related
memos without completing this evidence chain or checking every claim. ReAct
therefore reaches only 6\% at \$2.77, below direct inference at 44\% and \$1.17.

Across these examples, the value of additional inference depends on how it
organizes subsequent work. Targeted searches, reusable fact tables, and smaller
candidate sets support efficient verification; repeated reading and broad
candidate lists can add cost while leaving the core reasoning unresolved.

\begin{wraptable}{R}{0.40\linewidth}
  \centering
  \vspace{-0.7cm}
  \caption{GPT-5.6-sol on existing long-context benchmarks. Each cell reports
  accuracy and average USD per instance in parentheses.}
  \label{tab:existing-benchmarks}
  \scriptsize
  \setlength{\tabcolsep}{2.5pt}
  \begin{tabular}{@{}lcc@{}}
    \toprule
    Harness & OOLONG-Synth & LongBench-v2 \\
    \midrule
    Direct         & \accost{64}{0.686} & \accost{62}{0.532} \\
    RLM            & \accost{78}{3.70}  & \accost{62}{0.641} \\
    OpenCode       & \accost{74}{2.08}  & \accost{60}{0.463} \\
    mini-swe-agent & \accost{86}{0.217} & \accost{68}{0.190} \\
    ReAct          & \accost{70}{0.751} & \accost{68}{0.354} \\
    \bottomrule
  \end{tabular}
  \vspace{-0.5cm}
\end{wraptable}
\subsection{Comparison with Existing Long-Context Benchmarks}

Figure~\ref{fig:harness-task-effects} compares GPT-5.6-sol on our suite with
50 OOLONG-Synth 128K examples and a 50-example, 90K--128K LongBench-v2 subset
\citep{bertsch2025oolong,bai2024longbenchv2};
Table~\ref{tab:existing-benchmarks} gives their benchmark-wide averages.
Existing benchmarks produce a relatively simple picture of harness performance.
On OOLONG-Synth, every tested agentic harness improves over direct reading, with
accuracy ranging from 70\% to 86\% versus 64\% for direct reading. On LongBench-v2,
all five configurations lie within a narrow 60--68\% range. Despite cost
differences, these benchmarks show either uniformly positive accuracy gains or
weak separation between harnesses.

Across our four tasks, every harness both helps and hurts GPT-5.6-sol relative
to direct reading, spanning all four accuracy--cost regions in
Figure~\ref{fig:harness-task-effects}. This variation reveals which task demands
match a harness's processing strategy and when extra calls fail to simplify
the work.

The comparison also shows that context length alone does not determine
processing demand. Under a similar 128K-scale context budget, mini-swe-agent
reaches 68\% on our suite at \$0.816 per instance. It reaches the same accuracy
on LongBench-v2 at \$0.190 and 86\% on OOLONG-Synth at \$0.217. Exact success has
different meanings across the benchmarks, so these percentages are not a
task-normalized measure of difficulty. Nevertheless, the lower accuracy ceiling
and higher cost on our suite leave substantial room for future harnesses to
improve both effectiveness and efficiency.

%


\section{Related Work}

\paragraph{Long-context evaluation.}
Long Range Arena introduced a unified framework for evaluating models on long
inputs \citep{tay2021longrange}.  Subsequent benchmarks include broad suites
covering multiple tasks and context lengths
\citep{shaham2023zeroscrolls,an2024eval,bai2024longbench,
zhang2024infinitebench,yen2025helmet,bai2024longbenchv2}.
More targeted evaluations test retrieval and sensitivity to evidence position
\citep{liu2024lostmiddle,kamradt2023niah,hsieh2024ruler}, reasoning over
scattered evidence \citep{song2025countingstars,kuratov2024babilong}, and
aggregation across many records \citep{bertsch2025oolong,lin2025mebench}.
Others focus on following dependent procedural steps \citep{ye2025longproc}
or retrieving evidence that requires reasoning or distinguishing closely
related candidates \citep{su2025bright,modarressi2025nolima,luo2026muddle}.
These settings differ in how much evidence is needed and how it is distributed
throughout the context \citep{goldman2024longcontext}.  Our suite builds on
these challenges by requiring verification of retrieved evidence to guide
subsequent searches.  It also admits multiple solution strategies with
different computational costs, allowing us to evaluate both accuracy and
efficiency.

Each of our four tasks extends an existing evaluation setting.
QUEST retrieves sets of entities defined by implicit operations
\citep{malaviya2023quest}, while BrowseComp finds a single fact from several
clues \citep{wei2025browsecomp}.  Constraint Solving Search finds every entity
satisfying scattered conditions, with filtering order affecting computational
cost.  Logic Haystacks retrieves evidence for a specified contradiction
\citep{sileo2026logic}, and MAGIC judges a supplied context pair
\citep{lee2025magic}.  \Memos{} instead finds every conflict by combining
evidence across records.

EquiBench \citep{wei2025equibench} and SeqCoBench
\citep{maveli2025seqcobench} classify supplied program pairs.  Equivalent
Program Pair Search discovers all matching pairs while rejecting mutants with
similar code but different behavior.
CRUXEval predicts inputs or outputs for short standalone functions
\citep{gu2024cruxeval}, while RepoReasoner traces outputs and calls across
repository files \citep{wang2026reporeasoner}.  \Notebook{} traces dependent
computations by retrieving semantically equivalent functions and reusing
recorded outputs as subsequent inputs.  Across these tasks, verifying
plausible candidates guides what evidence to retrieve next.

\paragraph{LM harnesses as long-context processors.}
LM harnesses provide mechanisms for accessing and managing context across
multiple interactions.  ReAct interleaves reasoning with actions
\citep{yao2023react}, and SWE-agent uses search and execution to navigate
repositories \citep{yang2024sweagent}.  Other approaches manage the context
itself: RLM recursively decomposes prompts stored outside the model's context
window \citep{zhang2025rlm}, MemGPT manages memory across tiers
\citep{packer2023memgpt}, and Context as a Tool lets the model choose when to
compress its context \citep{liu2026contexttool}.  Studies of these processing
choices include LaRA, which compares retrieval with full-context processing
\citep{li2025lara}, and \citet{cao2026codingagents}, who evaluate coding agents
as general long-context processors.  Our suite provides a common setting for
comparing harness strategies while holding the underlying model and task
context fixed.

\paragraph{Efficiency-aware evaluation.}
Recent agent benchmarks evaluate both task success and resource use.
MLAgentBench and AgencyBench study these dimensions on long-horizon tasks
\citep{huang2024mlagentbench,li2026agencybench}.  The Holistic Agent Leaderboard
compares accuracy--cost trade-offs using Pareto frontiers \citep{kapoor2026hal}.
Studies of budget-aware reasoning and coding agents further show that using
more tokens does not necessarily improve accuracy
\citep{wang2024tokeneconomies,bai2026agentspend}.  We bring this efficiency-aware
perspective to long-context harness evaluation, measuring accuracy, token use,
and cost to test whether additional computation improves performance.

\section{Conclusion}

We introduced a suite for stress-testing long-context LM harnesses on four
information-dense tasks with deterministic evaluation.  Across five model
families, we find no universally best harness: accuracy gains depend on the
model and task, equally accurate systems can differ by more than an order of
magnitude in cost, and additional inference compute does not reliably improve
performance.  Long-context systems should therefore be evaluated as
model--harness combinations, with efficiency measured alongside accuracy.  Our
suite provides a testbed for developing harnesses that use additional compute
selectively and productively.



\bibliographystyle{iclr2027_conference}
\bibliography{references}

@inproceedings{
tay2021longrange,
title={Long Range Arena : A Benchmark for Efficient Transformers },
author={Yi Tay and Mostafa Dehghani and Samira Abnar and Yikang Shen and Dara Bahri and Philip Pham and Jinfeng Rao and Liu Yang and Sebastian Ruder and Donald Metzler},
booktitle={International Conference on Learning Representations},
year={2021},
url={https://openreview.net/forum?id=qVyeW-grC2k}
}

@inproceedings{bai2024longbench,
  title={Longbench: A bilingual, multitask benchmark for long context understanding},
  author={Bai, Yushi and Lv, Xin and Zhang, Jiajie and Lyu, Hongchang and Tang, Jiankai and Huang, Zhidian and Du, Zhengxiao and Liu, Xiao and Zeng, Aohan and Hou, Lei and others},
  booktitle={Proceedings of the 62nd annual meeting of the association for computational linguistics (volume 1: Long papers)},
  pages={3119--3137},
  year={2024}
}

@inproceedings{an2024eval,
  title={L-eval: Instituting standardized evaluation for long context language models},
  author={An, Chenxin and Gong, Shansan and Zhong, Ming and Zhao, Xingjian and Li, Mukai and Zhang, Jun and Kong, Lingpeng and Qiu, Xipeng},
  booktitle={Proceedings of the 62nd Annual Meeting of the Association for Computational Linguistics (Volume 1: Long Papers)},
  pages={14388--14411},
  year={2024}
}

@InProceedings{li2025lara,
  title = 	 {{L}a{RA}: Benchmarking Retrieval-Augmented Generation and Long-Context {LLM}s – No Silver Bullet for {LC} or {RAG} Routing},
  author =       {Li, Kuan and Zhang, Liwen and Jiang, Yong and Xie, Pengjun and Huang, Fei and Wang, Shuai and Cheng, Minhao},
  booktitle = 	 {Proceedings of the 42nd International Conference on Machine Learning},
  pages = 	 {36846--36867},
  year = 	 {2025},
  editor = 	 {Singh, Aarti and Fazel, Maryam and Hsu, Daniel and Lacoste-Julien, Simon and Berkenkamp, Felix and Maharaj, Tegan and Wagstaff, Kiri and Zhu, Jerry},
  volume = 	 {267},
  series = 	 {Proceedings of Machine Learning Research},
  month = 	 {13--19 Jul},
  publisher =    {PMLR},
  url = 	 {https://proceedings.mlr.press/v267/li25dv.html}
}

@inproceedings{wang2024tokeneconomies,
  title={Reasoning in token economies: Budget-aware evaluation of llm reasoning strategies},
  author={Wang, Junlin and Jain, Siddhartha and Zhang, Dejiao and Ray, Baishakhi and Kumar, Varun and Athiwaratkun, Ben},
  booktitle={Proceedings of the 2024 Conference on Empirical Methods in Natural Language Processing},
  pages={19916--19939},
  year={2024}
}

@inproceedings{li2026agencybench,
  title={Agencybench: Benchmarking the frontiers of autonomous agents in 1m-token real-world contexts},
  author={Li, Keyu and Shi, Junhao and Xiao, Yang and Jiang, Mohan and Sun, Jie and Wu, Yunze and Fu, Dayuan and Xia, Shijie and Cai, Xiaojie and Xu, Tianze and others},
  booktitle={Proceedings of the 64th Annual Meeting of the Association for Computational Linguistics (Volume 1: Long Papers)},
  pages={7422--7440},
  year={2026}
}

@article{packer2023memgpt,
  title={Memgpt: Towards llms as operating systems},
  author={Packer, Charles and Wooders, Sarah and Lin, Kevin and Fang, Vivian and Patil, Shishir G and Stoica, Ion and Gonzalez, Joseph E},
  journal={arXiv preprint arXiv:2310.08560},
  year={2023}
}

@inproceedings{su2025bright,
  title={Bright: A realistic and challenging benchmark for reasoning-intensive retrieval},
  author={Su, Hongjin and Yen, Howard and Xia, Mengzhou and Shi, Weijia and Muennighoff, Niklas and Wang, Han-yu and Haisu, Liu and Shi, Quan and Siegel, Zachary and Tang, Michael and others},
  booktitle={International Conference on Learning Representations},
  volume={2025},
  pages={48941--48991},
  year={2025}
}

@InProceedings{gu2024cruxeval,
  title = 	 {{CRUXE}val: A Benchmark for Code Reasoning, Understanding and Execution},
  author =       {Gu, Alex and Roziere, Baptiste and Leather, Hugh James and Solar-Lezama, Armando and Synnaeve, Gabriel and Wang, Sida},
  booktitle = 	 {Proceedings of the 41st International Conference on Machine Learning},
  pages = 	 {16568--16621},
  year = 	 {2024},
  editor = 	 {Salakhutdinov, Ruslan and Kolter, Zico and Heller, Katherine and Weller, Adrian and Oliver, Nuria and Scarlett, Jonathan and Berkenkamp, Felix},
  volume = 	 {235},
  series = 	 {Proceedings of Machine Learning Research},
  month = 	 {21--27 Jul},
  publisher =    {PMLR},
  url = 	 {https://proceedings.mlr.press/v235/gu24c.html}
}

@inproceedings{bai2024longbenchv2,
  title={Longbench v2: Towards deeper understanding and reasoning on realistic long-context multitasks},
  author={Bai, Yushi and Tu, Shangqing and Zhang, Jiajie and Peng, Hao and Wang, Xiaozhi and Lv, Xin and Cao, Shulin and Xu, Jiazheng and Hou, Lei and Dong, Yuxiao and others},
  booktitle={Proceedings of the 63rd Annual Meeting of the Association for Computational Linguistics (Volume 1: Long Papers)},
  pages={3639--3664},
  year={2025}
}

@article{bertsch2025oolong,
  title={Oolong: Evaluating long context reasoning and aggregation capabilities},
  author={Bertsch, Amanda and Pratapa, Adithya and Mitamura, Teruko and Neubig, Graham and Gormley, Matthew R},
  journal={arXiv preprint arXiv:2511.02817},
  year={2025}
}

@article{wei2025browsecomp,
  title={Browsecomp: A simple yet challenging benchmark for browsing agents},
  author={Wei, Jason and Sun, Zhiqing and Papay, Spencer and McKinney, Scott and Han, Jeffrey and Fulford, Isa and Chung, Hyung Won and Passos, Alex Tachard and Fedus, William and Glaese, Amelia},
  journal={arXiv preprint arXiv:2504.12516},
  year={2025}
}

@article{yang2024sweagent,
  title={Swe-agent: Agent-computer interfaces enable automated software engineering},
  author={Yang, John and Jimenez, Carlos and Wettig, Alexander and Lieret, Kilian and Yao, Shunyu and Narasimhan, Karthik and Press, Ofir},
  journal={Advances in Neural Information Processing Systems},
  volume={37},
  pages={50528--50652},
  year={2024}
}

@article{zhang2025rlm,
  title={Recursive language models},
  author={Zhang, Alex L and Kraska, Tim and Khattab, Omar},
  journal={arXiv preprint arXiv:2512.24601},
  year={2025}
}

@inproceedings{
yao2023react,
title={ReAct: Synergizing Reasoning and Acting in Language Models},
author={Shunyu Yao and Jeffrey Zhao and Dian Yu and Nan Du and Izhak Shafran and Karthik R Narasimhan and Yuan Cao},
booktitle={The Eleventh International Conference on Learning Representations },
year={2023},
url={https://openreview.net/forum?id=WE_vluYUL-X}
}

@article{cao2026codingagents,
  title={Coding agents are effective long-context processors},
  author={Cao, Weili and Yin, Xunjian and Dhingra, Bhuwan and Zhou, Shuyan},
  journal={arXiv preprint arXiv:2603.20432},
  year={2026}
}

@inproceedings{
hsieh2024ruler,
title={{RULER}: What{\textquoteright}s the Real Context Size of Your Long-Context Language Models?},
author={Cheng-Ping Hsieh and Simeng Sun and Samuel Kriman and Shantanu Acharya and Dima Rekesh and Fei Jia and Boris Ginsburg},
booktitle={First Conference on Language Modeling},
year={2024},
url={https://openreview.net/forum?id=kIoBbc76Sy}
}

@inproceedings{
ye2025longproc,
title={LongProc: Benchmarking Long-Context Language Models on Long Procedural Generation},
author={Xi Ye and Fangcong Yin and Yinghui He and Joie Zhang and Howard Yen and Tianyu Gao and Greg Durrett and Danqi Chen},
booktitle={Second Conference on Language Modeling},
year={2025},
url={https://openreview.net/forum?id=ruWC5LIMSo}
}

@inproceedings{shaham2023zeroscrolls,
  title={ZeroSCROLLS: A zero-shot benchmark for long text understanding},
  author={Shaham, Uri and Ivgi, Maor and Efrat, Avia and Berant, Jonathan and Levy, Omer},
  booktitle={Findings of the Association for Computational Linguistics: EMNLP 2023},
  pages={7977--7989},
  year={2023}
}

@inproceedings{zhang2024infinitebench,
    title = "$\infty${B}ench: Extending Long Context Evaluation Beyond 100{K} Tokens",
    author = "Zhang, Xinrong  and
      Chen, Yingfa  and
      Hu, Shengding  and
      Xu, Zihang  and
      Chen, Junhao  and
      Hao, Moo  and
      Han, Xu  and
      Thai, Zhen  and
      Wang, Shuo  and
      Liu, Zhiyuan  and
      Sun, Maosong",
    editor = "Ku, Lun-Wei  and
      Martins, Andre  and
      Srikumar, Vivek",
    booktitle = "Proceedings of the 62nd Annual Meeting of the Association for Computational Linguistics (Volume 1: Long Papers)",
    month = aug,
    year = "2024",
    address = "Bangkok, Thailand",
    publisher = "Association for Computational Linguistics",
    url = "https://aclanthology.org/2024.acl-long.814/",
    doi = "10.18653/v1/2024.acl-long.814",
    pages = "15262--15277"
}

@inproceedings{
yen2025helmet,
title={{HELMET}: How to Evaluate Long-context Models Effectively and Thoroughly},
author={Howard Yen and Tianyu Gao and Minmin Hou and Ke Ding and Daniel Fleischer and Peter Izsak and Moshe Wasserblat and Danqi Chen},
booktitle={The Thirteenth International Conference on Learning Representations},
year={2025},
url={https://openreview.net/forum?id=293V3bJbmE}
}

@misc{opencode2026,
  title        = {{OpenCode}},
  author       = {{Anomaly}},
  howpublished = {\url{https://opencode.ai}},
  note         = {Version 1.18.21},
  year         = {2026}
}

@misc{openai2026gpt56sol,
  title        = {{GPT-5.6 Sol} Model},
  author       = {{OpenAI}},
  howpublished = {\url{https://developers.openai.com/api/docs/models/gpt-5.6-sol}},
  year         = {2026}
}

@article{glm5team2026glm5,
  title={Glm-5: from vibe coding to agentic engineering},
  author={Zeng, Aohan and Lv, Xin and Hou, Zhenyu and Du, Zhengxiao and Zheng, Qinkai and Chen, Bin and Yin, Da and Ge, Chendi and Huang, Chenghua and Xie, Chengxing and others},
  journal={arXiv preprint arXiv:2602.15763},
  year={2026}
}

@misc{qwen2026qwen38,
    title = {{Qwen3.8-Max}: A New Bar for Coding and Cowork},
    url = {https://qwen.ai/blog?id=qwen3.8},
    author = {{Qwen Team}},
    month = {August},
    year = {2026}
}

@misc{moonshot2026kimik26,
  title        = {{Kimi-K2.6}},
  author       = {{Moonshot AI}},
  howpublished = {\url{https://huggingface.co/moonshotai/Kimi-K2.6}},
  year         = {2026}
}

@inproceedings{
hendrycks2021apps,
title={Measuring Coding Challenge Competence With {APPS}},
author={Dan Hendrycks and Steven Basart and Saurav Kadavath and Mantas Mazeika and Akul Arora and Ethan Guo and Collin Burns and Samir Puranik and Horace He and Dawn Song and Jacob Steinhardt},
booktitle={Thirty-fifth Conference on Neural Information Processing Systems Datasets and Benchmarks Track (Round 2)},
year={2021},
url={https://openreview.net/forum?id=sD93GOzH3i5}
}

@article{liu2024lostmiddle,
  title={Lost in the middle: How language models use long contexts},
  author={Liu, Nelson F and Lin, Kevin and Hewitt, John and Paranjape, Ashwin and Bevilacqua, Michele and Petroni, Fabio and Liang, Percy},
  journal={Transactions of the association for computational linguistics},
  volume={12},
  pages={157--173},
  year={2024}
}

@article{kuratov2024babilong,
  title={Babilong: Testing the limits of llms with long context reasoning-in-a-haystack},
  author={Kuratov, Yuri and Bulatov, Aydar and Anokhin, Petr and Rodkin, Ivan and Sorokin, Dmitry and Sorokin, Artyom and Burtsev, Mikhail},
  journal={Advances in Neural Information Processing Systems},
  volume={37},
  pages={106519--106554},
  year={2024}
}

@inproceedings{
modarressi2025nolima,
title={NoLiMa: Long-Context Evaluation Beyond Literal Matching},
author={Ali Modarressi and Hanieh Deilamsalehy and Franck Dernoncourt and Trung Bui and Ryan A. Rossi and Seunghyun Yoon and Hinrich Schuetze},
booktitle={Forty-second International Conference on Machine Learning},
year={2025},
url={https://openreview.net/forum?id=0OshX1hiSa}
}

@inproceedings{sileo2026logic,
  title={Logic Haystacks: Probing LLMs’ Long-Context Logical Reasoning (Without Easily Identifiable Unrelated Padding)},
  author={Sileo, Damien},
  booktitle={Proceedings of the 19th Conference of the European Chapter of the Association for Computational Linguistics (Volume 2: Short Papers)},
  pages={66--75},
  year={2026}
}

@inproceedings{lee2025magic,
    title = "{MAGIC}: A Multi-Hop and Graph-Based Benchmark for Inter-Context Conflicts in Retrieval-Augmented Generation",
    author = "Lee, Jungyeon  and
      Kangmin, Lee  and
      Kim, Taeuk",
    editor = "Christodoulopoulos, Christos  and
      Chakraborty, Tanmoy  and
      Rose, Carolyn  and
      Peng, Violet",
    booktitle = "Findings of the Association for Computational Linguistics: EMNLP 2025",
    month = nov,
    year = "2025",
    address = "Suzhou, China",
    publisher = "Association for Computational Linguistics",
    url = "https://aclanthology.org/2025.findings-emnlp.466/",
    doi = "10.18653/v1/2025.findings-emnlp.466",
    pages = "8783--8803",
    ISBN = "979-8-89176-335-7"
}

@inproceedings{malaviya2023quest,
  title={Quest: A retrieval dataset of entity-seeking queries with implicit set operations},
  author={Malaviya, Chaitanya and Shaw, Peter and Chang, Ming-Wei and Lee, Kenton and Toutanova, Kristina},
  booktitle={Proceedings of the 61st Annual Meeting of the Association for Computational Linguistics (Volume 1: Long Papers)},
  pages={14032--14047},
  year={2023}
}

@inproceedings{wei2025equibench,
    title = "{E}qui{B}ench: Benchmarking Large Language Models' Reasoning about Program Semantics via Equivalence Checking",
    author = "Wei, Anjiang  and
      Cao, Jiannan  and
      Li, Ran  and
      Chen, Hongyu  and
      Zhang, Yuhui  and
      Wang, Ziheng  and
      Liu, Yuan  and
      Teixeira, Thiago S. F. X.  and
      Yang, Diyi  and
      Wang, Ke  and
      Aiken, Alex",
    editor = "Christodoulopoulos, Christos  and
      Chakraborty, Tanmoy  and
      Rose, Carolyn  and
      Peng, Violet",
    booktitle = "Proceedings of the 2025 Conference on Empirical Methods in Natural Language Processing",
    month = nov,
    year = "2025",
    address = "Suzhou, China",
    publisher = "Association for Computational Linguistics",
    url = "https://aclanthology.org/2025.emnlp-main.1718/",
    doi = "10.18653/v1/2025.emnlp-main.1718",
    pages = "33868--33881",
    ISBN = "979-8-89176-332-6"
}

@inproceedings{kapoor2026hal,
  title={Holistic agent leaderboard: The missing infrastructure for ai agent evaluation},
  author={Kapoor, Sayash and Stroebl, Benedikt and Kirgis, Peter and Nadgir, Nitya and Siegel, Zachary and Wei, Boyi and Xue, Tianci and Chen, Ziru and Chen, Felix and Utpala, Saiteja and others},
  booktitle={International Conference on Learning Representations},
  volume={2026},
  pages={98778--98849},
  year={2026}
}

@article{bai2026agentspend,
  title={How do ai agents spend your money? analyzing and predicting token consumption in agentic coding tasks},
  author={Bai, Longju and Huang, Zhemin and Wang, Xingyao and Sun, Jiao and Mihalcea, Rada and Brynjolfsson, Erik and Pentland, Alex and Pei, Jiaxin},
  journal={arXiv preprint arXiv:2604.22750},
  year={2026}
}

@inproceedings{goldman2024longcontext,
  title={Is it really long context if all you need is retrieval? towards genuinely difficult long context nlp},
  author={Goldman, Omer and Jacovi, Alon and Slobodkin, Aviv and Maimon, Aviya and Dagan, Ido and Tsarfaty, Reut},
  booktitle={Proceedings of the 2024 Conference on Empirical Methods in Natural Language Processing},
  pages={16576--16586},
  year={2024}
}

@inproceedings{song2025countingstars,
  title={Counting-stars: A multi-evidence, position-aware, and scalable benchmark for evaluating long-context large language models},
  author={Song, Mingyang and Zheng, Mao and Luo, Xuan},
  booktitle={Proceedings of the 31st International Conference on Computational Linguistics},
  pages={3753--3763},
  year={2025}
}

@inproceedings{lin2025mebench,
  title={Mebench: Benchmarking large language models for cross-document multi-entity question answering},
  author={Lin, Teng and Luo, Yuyu and Zhang, Honglin and Zhang, Jicheng and Liu, Chunlin and Wu, Kaishun and Tang, Nan},
  booktitle={Proceedings of the 2025 Conference on Empirical Methods in Natural Language Processing},
  pages={1481--1494},
  year={2025}
}

@inproceedings{
luo2026muddle,
title={{MUDDLE}: Measuring Understanding of Documents under Distractor and Length Effects},
author={Jason Luo and Saibilila Abudukelimu and Judy Song and Andrew Feng and Shivank Garg and Vasu Sharma and Kevin Zhu},
booktitle={COLM Workshop on Context Beyond the Window: Persistent Knowledge in Language Models},
year={2026},
url={https://openreview.net/forum?id=9YS7tSPM9c}
}

@inproceedings{maveli2025seqcobench,
  title={What can large language models capture about code functional equivalence?},
  author={Maveli, Nickil and Vergari, Antonio and Cohen, Shay B},
  booktitle={Findings of the Association for Computational Linguistics: NAACL 2025},
  pages={6880--6918},
  year={2025}
}

@article{wang2026reporeasoner,
  title={RepoReasoner: Evaluating Repository-Level Code Reasoning Ability of Long-Context Language Models},
  author={Wang, Yanlin and Wang, Suiquan and Wang, Yanli and Zhang, Bowen and Guo, Daya and Chen, Jiachi and Zheng, Zibin},
  journal={Proceedings of the ACM on Software Engineering},
  volume={3},
  number={FSE},
  pages={2790--2812},
  year={2026},
  publisher={ACM New York, NY, USA}
}

@inproceedings{liu2026contexttool,
  title={Context as a tool: Context management for long-horizon swe-agents},
  author={Liu, Shukai and Jiang, Bo and Yang, Jian and Li, Yizhi and Guo, Jinyang and Liu, Xianglong and Dai, Bryan},
  booktitle={Findings of the Association for Computational Linguistics: ACL 2026},
  pages={20604--20617},
  year={2026}
}

@InProceedings{huang2024mlagentbench,
  title = 	 {{MLA}gent{B}ench: Evaluating Language Agents on Machine Learning Experimentation},
  author =       {Huang, Qian and Vora, Jian and Liang, Percy and Leskovec, Jure},
  booktitle = 	 {Proceedings of the 41st International Conference on Machine Learning},
  pages = 	 {20271--20309},
  year = 	 {2024},
  editor = 	 {Salakhutdinov, Ruslan and Kolter, Zico and Heller, Katherine and Weller, Adrian and Oliver, Nuria and Scarlett, Jonathan and Berkenkamp, Felix},
  volume = 	 {235},
  series = 	 {Proceedings of Machine Learning Research},
  month = 	 {21--27 Jul},
  publisher =    {PMLR},
  url = 	 {https://proceedings.mlr.press/v235/huang24y.html}
}

@article{huang2026memoharness,
  title={MemoHarness: Agent Harnesses That Learn from Experience},
  author={Huang, Yue and Wang, Wenjie and Bao, Han and Ma, Yuchen and Luo, Xiaonan and Nian, Yi and Zhuang, Haomin and Liu, Zheyuan and Zhao, Yue and Zhang, Xiangliang},
  journal={arXiv preprint arXiv:2607.14159},
  year={2026}
}

@inproceedings{bge-m3,
  title={M3-embedding: Multi-linguality, multi-functionality, multi-granularity text embeddings through self-knowledge distillation},
  author={Chen, Jianlyu and Xiao, Shitao and Zhang, Peitian and Luo, Kun and Lian, Defu and Liu, Zheng},
  booktitle={Findings of the association for computational linguistics: ACL 2024},
  pages={2318--2335},
  year={2024}
}

@misc{kamradt2023niah,
  author = {Greg Kamradt},
  title = {Needle In A Haystack - Pressure Testing LLMs},
  year = {2023},
  publisher = {GitHub},
  journal = {GitHub repository},
  howpublished = {\url{https://github.com/gkamradt/needle-in-a-haystack}}
}

\appendix
\clearpage
\section{Detailed Results}
\label{app:detailed-results}
\begin{table}[H]
  \caption{Detailed accuracy and efficiency on Constraint Solving Search (Constraint),
  Outlier Memo Detection (Memos), Program Execution Tracing (Program), and
  Equivalent Program Pair Search (Pairs).  Macro accuracy averages the four
  task-level accuracies. Program uses the format-tolerant answer-and-citation
  rescore; its original strict-format scores remain in the task report.}
  \label{tab:detailed-results}
  \centering
  \scriptsize
  \newcolumntype{Y}{>{\raggedleft\arraybackslash}X}
  \textbf{(a) Accuracy (\%)}\\[2pt]
  \begin{tabularx}{\textwidth}{@{}ll*{5}{Y}@{}}
    \toprule
    Model & Harness & Constraint & Memos & Program & Pairs & Macro avg. \\
    \midrule
    GPT-5.6-sol & Direct         & 58\% & 44\% & 94\% & 44\% & 60.0\% \\
                & RLM            & 44\% & 92\% & 76\% &  2\% & 53.5\% \\
                & OpenCode       & 56\% & 10\% & 96\% &  0\% & 40.5\% \\
                & mini-swe-agent & 42\% & 100\% & 88\% & 42\% & 68.0\% \\
                & ReAct          & 32\% &  6\% & 82\% & 52\% & 43.0\% \\
    \midrule
    GLM-5.3 & Direct         & 0\% &  0\% &  38\% & 0\% &  9.5\% \\
            & RLM            & 34\% & 16\% & 58\% & 0\% & 27.0\% \\
            & OpenCode       & 48\% & 20\% & 84\% & 0\% & 38.0\% \\
            & mini-swe-agent & 66\% & 58\% & 46\% & 0\% & 42.5\% \\
            & ReAct          & 14\% &  0\% & 34\% & 4\% & 13.0\% \\
    \midrule
    Qwen3.8-27B & Direct         &  0\% & 0\% &  0\% & 0\% &  0.0\% \\
                 & RLM            & 14\% & 2\% & 14\% & 0\% &  7.5\% \\
                 & OpenCode       & 46\% & 16\% & 92\% & 0\% & 38.5\% \\
                 & mini-swe-agent & 52\% & 14\% & 66\% & 0\% & 33.0\% \\
                 & ReAct          &  0\% &  0\% &  0\% & 0\% &  0.0\% \\
    \midrule
    Gemini 3.8 Flash & Direct         & 12\% &  0\% & 28\% & 0\% & 10.0\% \\
                     & RLM            &  20\% & 22\% & 0\% & 0\% & 10.5\% \\
                     & OpenCode       & 0\% & 28\% &  18\% & 0\% & 11.5\% \\
                     & mini-swe-agent & 4\% & 64\% &  32\% & 0\% & 25.0\% \\
                     & ReAct          & 0\% & 30\% &  26\% & 28\% & 21.0\% \\
    \midrule
    Kimi-K2.6 & Direct         & 0\% & 0\% & 0\% & 0\% & 0.0\% \\
              & RLM            & 2\% & 0\% & 0\% & 0\% & 0.5\% \\
              & OpenCode       & 0\% & 0\% & 6\% & 0\% & 1.5\% \\
              & mini-swe-agent & 0\% & 0\% & 2\% & 0\% & 0.5\% \\
              & ReAct          & 0\% & 0\% & 0\% & 0\% & 0.0\% \\
    \bottomrule
  \end{tabularx}
  \vspace{4pt}
  \textbf{(b) Efficiency: average input / output tokens (millions) and cost per instance}\\[2pt]
  \setlength{\tabcolsep}{1pt}
  \begin{tabularx}{\textwidth}{@{}ll*{12}{Y}@{}}
    \toprule
    & & \multicolumn{3}{c}{Constraint} & \multicolumn{3}{c}{Memos}
      & \multicolumn{3}{c}{Program} & \multicolumn{3}{c}{Pairs} \\
    \cmidrule(lr){3-5}\cmidrule(lr){6-8}\cmidrule(lr){9-11}\cmidrule(l){12-14}
    Model & Harness & Input & Output & Cost & Input & Output & Cost
      & Input & Output & Cost & Input & Output & Cost \\
    \midrule
    GPT-5.6-sol & Direct & 0.107 & 0.0077 & \$0.690 & 0.0646 & 0.0440 & \$1.17 & 0.121 & 0.0026 & \$0.648 & 0.0818 & 0.0185 & \$0.780 \\
                 & RLM & 0.250 & 0.0180 & \$0.659 & 0.745 & 0.283 & \$7.23 & 0.404 & 0.0285 & \$1.47 & 0.712 & 0.274 & \$7.86 \\
                 & OpenCode & 1.80 & 0.0058 & \$1.72 & 2.49 & 0.0241 & \$3.11 & 1.29 & 0.0045 & \$1.37 & 2.75 & 0.0606 & \$5.41 \\
                 & mini-swe-agent & 0.181 & 0.0070 & \$0.320 & 0.367 & 0.0133 & \$0.583 & 1.41 & 0.0058 & \$1.02 & 1.47 & 0.0223 & \$1.34 \\
                 & ReAct & 1.28 & 0.0068 & \$1.40 & 2.54 & 0.0211 & \$2.77 & 0.636 & 0.0035 & \$0.866 & 1.43 & 0.0217 & \$2.68 \\
    \midrule
    GLM-5.3 & Direct & 0.110 & 0.0161 & \$0.728 & 0.0664 & 0.0640 & \$1.09 & 0.121 & 0.0170 & \$0.785 & 0.0818$^{\dagger}$ & 0.0655$^{\dagger}$ & \$1.19$^{\dagger}$ \\
             & RLM & 1.76 & 0.0524 & \$2.67 & 1.49 & 0.488 & \$7.84 & 2.02 & 0.0583 & \$3.34 & 2.02 & 0.384 & \$8.06 \\
             & OpenCode & 1.20 & 0.0307 & \$1.94 & 2.35 & 0.0568 & \$3.52 & 2.80 & 0.0555 & \$4.22 & 2.28 & 0.0700 & \$3.64 \\
             & mini-swe-agent & 1.09 & 0.0227 & \$1.55 & 1.91 & 0.0611 & \$3.00 & 3.21 & 0.0154 & \$3.65 & 2.64 & 0.0648 & \$3.80 \\
             & ReAct & 1.81 & 0.0360 & \$2.78 & 1.71 & 0.133 & \$4.12 & 1.73 & 0.0267 & \$2.59 & 1.47 & 0.0768 & \$3.26 \\
    \midrule
    Qwen3.8-27B & Direct & 0.120 & 0.0077 & \$0.356 & 0.0704 & 0.0158 & \$0.292 & 0.133 & 0.0046 & \$0.363 & 0.0887 & 0.0162 & \$0.341 \\
                  & RLM & 0.927 & 0.0272 & \$2.50 & 2.27 & 0.215 & \$7.23 & 2.09 & 0.0794 & \$5.78 & 1.68 & 0.398 & \$7.13 \\
                  & OpenCode & 2.48 & 0.0455 & \$6.50 & 2.38 & 0.0575 & \$6.32 & 3.30 & 0.0542 & \$8.59 & 2.58 & 0.0563 & \$6.83 \\
                  & mini-swe-agent & 0.968 & 0.0272 & \$2.60 & 2.28 & 0.119 & \$6.54 & 2.07 & 0.0237 & \$5.30 & 2.66 & 0.0330 & \$6.84 \\
                  & ReAct & 1.27 & 0.0294 & \$3.37 & 1.80 & 0.198 & \$5.93 & 0.501 & 0.0737 & \$1.79 & 1.43 & 0.114 & \$4.39 \\
    \midrule
    Gemini 3.8 Flash & Direct & 0.116 & 0.0236 & \$0.166 & 0.0679 & 0.0293 & \$0.158 & 0.137 & 0.0201 & \$0.170 & 0.0961 & 0.0257 & \$0.154 \\
                      & RLM & 2.51 & 0.0275 & \$0.505 & 2.65 & 0.0737 & \$0.729 & 2.56 & 0.0251 & \$0.478 & 2.66 & 0.168 & \$1.14 \\
                      & OpenCode & 2.58 & 0.0174 & \$0.439 & 2.52 & 0.0395 & \$0.515 & 2.59 & 0.0318 & \$0.478 & 2.78 & 0.0234 & \$0.452 \\
                      & mini-swe-agent & 2.74 & 0.0415 & \$0.611 & 2.23 & 0.0497 & \$0.528 & 2.62 & 0.0372 & \$0.507 & 2.81 & 0.0415 & \$0.554 \\
                      & ReAct & 2.75 & 0.0137 & \$0.428 & 2.10 & 0.102 & \$0.764 & 2.54 & 0.0216 & \$0.428 & 1.80 & 0.0491 & \$0.538 \\
    \midrule
    Kimi-K2.6 & Direct & 0.110 & 0.0080 & \$0.670 & 0.0653 & 0.0164 & \$0.546 & 0.120 & 0.0060 & \$0.696 & 0.0819 & 0.0164 & \$0.631 \\
               & RLM & 1.42 & 0.0491 & \$2.41 & 2.21 & 1.04 & \$16.3 & 2.20 & 0.0967 & \$3.92 & 1.48 & 0.143 & \$3.96 \\
               & OpenCode & 2.51 & 0.0377 & \$3.86 & 2.67 & 0.123 & \$5.64 & 5.83 & 0.0987 & \$8.99 & 2.65 & 0.0474 & \$4.10 \\
               & mini-swe-agent & 2.46 & 0.0273 & \$3.15 & 1.99 & 0.132 & \$4.22 & 4.30 & 0.0242 & \$5.08 & 2.59 & 0.0509 & \$3.65 \\
               & ReAct & 2.39 & 0.0196 & \$3.14 & 1.82 & 0.0222 & \$2.72 & 2.48 & 0.0287 & \$3.42 & 1.36 & 0.0171 & \$2.16 \\
    \bottomrule
  \end{tabularx}
\end{table}
\clearpage

\section{Dataset Construction Details}
\label{app:construction}

\paragraph{Common construction principle.}
For each instance, we first construct the underlying task as structured data.
This consists of predicate support sets for Constraint Solving Search, a typed
relation graph for Outlier Memo Detection, a computation graph for Program
Execution Tracing, or executable program families for Equivalent Program Pair
Search.  These hidden records determine the answer before the public context is
written.  A task-specific solver or execution engine computes the gold answer
and its supporting evidence.  We then create hard distractors by making
controlled changes to the records, render all items into their public form, and
shuffle them.  Finally, deterministic, rule-based checks recompute the answer
from the same records and verify that every required item and distractor
satisfies its intended role.

\begin{table}[h!]
  \caption{Underlying structures used to construct the four tasks.  Public
  contexts expose only the rendered items, not their roles or provenance.}
  \label{tab:datasets}
  \centering
  \footnotesize
  \begin{tabular}{@{}>{\raggedright\arraybackslash}p{0.20\linewidth}
                      >{\raggedright\arraybackslash}p{0.23\linewidth}
                      >{\raggedright\arraybackslash}p{0.26\linewidth}
                      >{\raggedright\arraybackslash}p{0.18\linewidth}@{}}
    \toprule
    Task & Private specification & Controlled distractors & Gold derivation \\
    \midrule
    Constraint Solving Search
      & Typed fact world and predicate support sets
      & People satisfying all but one condition
      & Set intersection and fact-to-passage map \\
    Outlier Memo Detection
      & Typed relation graph and composed relations
      & Endpoint substitutions and near-twin claims
      & Graph-composition proofs \\
    Program Execution Tracing
      & Computation graph, typed inputs, and function contracts
      & Same-output functions and alternate routes
      & Execution and contract probes \\
    Equivalent Program Pair Search
      & Executable program families and test signatures
      & Behavior-changing syntax-tree mutations
      & Differential execution \\
    \bottomrule
  \end{tabular}
\end{table}

\subsection{Constraint Solving Search}
\label{app:constraints-construction}

We first create a typed world containing 160 people and related entities, then
sample three to five predicates for the query.  Predicate templates cover
habits, current states, roles, counts, comparisons, and relations through
other entities.  We construct their support sets so that exactly five people
belong to every set.  For each predicate, two to four additional people satisfy
all other predicates but fail that one.  We also require one predicate to hold
for at most 10\% of the people and another to hold for at least 60\%.  This
symbolic set construction fixes the answer while deliberately making the best
search order much cheaper than checking every person against every condition.

We next instantiate the support sets as typed facts.  A failed predicate is
represented by a close but invalid fact, such as a planned bus trip rather
than a transit habit, instead of by a missing mention.  Facts are assigned
stable IDs, placed into passage plans of two to five statements, and mixed with
unrelated background facts.  The query, answer set, and near-match labels are
not part of the passage plans.  We retain every passage containing decisive
evidence, add neutral passages until the context reaches 120K--128K tokens,
and shuffle the passages.

The private solver intersects the predicate support sets and maps each matched
person and condition to the passages that express its supporting facts.  The
validator independently resolves the world, checks that every near-match fails
exactly one predicate, enforces the selective and broad support sizes, and
verifies that every cited fact survives packing.

\subsection{Outlier Memo Detection}
\label{app:memos-construction}

We construct each instance as a typed relation graph.  A two-hop rule composes
edges $R_1(a,b)$ and $R_2(b,c)$ into the claim $R_2\!\circ\!R_1(a,c)$.  The
generator samples 450 primitive edges and computes 440 such derived claims.
It then selects 15 derived claims and replaces the correct endpoint $c$ with a
different entity $c'$ of the same type.  The two primitive edges establishing
the correct endpoint remain elsewhere in the packet, so each corruption has a
complete symbolic contradiction proof.  Near-twin decoys reuse the same
relation types, and a false endpoint may be correct for a different anchor or
relation, preventing an entity-name shortcut.

We realize primitive and derived facts with multiple relation-specific
paraphrases, then pack three or four unrelated statements into each memo.  The
location of a corrupted statement within a memo is randomized, and target,
support, decoy, and background memos are matched in length.  The final packet
contains 700 memos and exactly 2,500 atomic statements drawn from five
operational domains.

For every target, the private record stores the corrupted claim, its correct
endpoint, the two composing edges, and the memo IDs containing their evidence.
Validation recomputes every composition from the relation graph, confirms that
exactly 15 published claims have incorrect endpoints, and checks that the
frozen instances regenerate byte for byte.

\subsection{Program Execution Tracing}
\label{app:notebook-construction}

We first sample an eight-node computation graph and assign each node a typed
transformation.  The transformations come from ten families covering table and
set operations, sequences and text, intervals, grids, paths, graph traversal,
and topological batching.  We execute this graph to obtain the input and output
at every node; an earlier output may become a later input, choose a branch, or
be reused by several nodes.  Across instances we use a linear chain and five
nonlinear structures: a diamond, skip routing, output reuse, multi-parent
routing, and a nested side chain.

For each node, we construct 20 cells evaluated on the input selected by the
gold trace.  Exactly one implements the node's transformation over its full
input contract.  The other 19 contain controlled changes to boundary handling,
filtering, aggregation, or ordering.  They are retained only when they produce
the same recorded output on the displayed call but disagree on a separate
contract probe.  We construct another 20-cell pool on an alternate input that
would arise after an incorrect earlier choice.  The pools are not marked in
the public context: all $8\times40=320$ cells are assigned anonymous IDs and
shuffled.  Each cell exposes only its function, example call, and recorded
output.

The answer key is obtained by executing the sampled graph and following its
selected route.  Validation reruns every cell, checks the function contracts
over generated probes, verifies that each query node has exactly one equivalent
cell on the gold input, and confirms that the cited cell outputs assemble into
the recorded final JSON.  This is behavioral validation over explicit input
contracts rather than a proof of equivalence for arbitrary Python objects.

\subsection{Equivalent Program Pair Search}
\label{app:pairs-construction}

We begin with standard-input APPS problems \citep{hendrycks2021apps}.  For each
problem, we retain four structurally distinct Python solutions that produce the
reference outputs on all supplied tests.  We then apply controlled syntax-tree
edits to each source style.  A mutant is retained only if it terminates on the
test suite, differs from the reference on at least one test, and has an output
signature distinct from the other retained mutants.  The two implementations
chosen as a gold pair must also differ in control-flow profile.  After
additional valid-input checks and manual review, families with disagreements or
ambiguous output contracts are removed; the evaluated suite draws from 76
retained families.

Each instance contains ten 20-program neighborhoods.  Five neighborhoods
contain two unmodified passing programs and 18 mutants, creating one gold pair
each.  The other five contain 20 mutants and no gold pair.  Both kinds contain
mutants from all four source styles, so identifying a common coding style does
not reveal whether a gold pair exists.  We further require each true mate to
rank below at least twelve mutants under a similarity measure combining
normalized tokens, syntax-tree features, and character n-grams.  All 200
programs are assigned anonymous IDs and shuffled; problem statements, source
identifiers, mutation labels, and input--output examples are omitted.

The answer key is the five pairs of unmodified programs.  Independent
validation replays every program on the supplied tests under multiple hash
seeds, checks the positive and negative output signatures, verifies the
neighborhood composition and similarity constraint, and recounts the
80K--120K-token context.  These are operational equivalence labels supported
by tests and additional probes, not formal proofs over every valid input.

\subsection{Existing Benchmark Subsets}
\label{app:existing-benchmark-subsets}

\paragraph{OOLONG-Synth.}
For OOLONG-Synth \citep{bertsch2025oolong}, we first retain the 400 test
examples in the official \texttt{context\_len}$=131{,}072$ condition.  We then
construct two disjoint 25-example cohorts using fixed random seeds.  Within
each cohort, we divide the examples as evenly as possible among the counting,
timeline, and user task groups, cycle across task types within each group, and
prefer source datasets and context-window IDs not yet represented.  The second
cohort excludes every task ID selected for the first.  The resulting 50-example
subset contains 17 counting, 16 timeline, and 17 user examples; each cohort
covers all eight source datasets and all 16 context-window IDs.  We use the
standard unlabeled input, which concatenates \texttt{context\_window\_text}
with the question, and exclude the label-annotated context provided by the
dataset.

\paragraph{LongBench-v2.}
For LongBench-v2 \citep{bai2024longbenchv2}, we tokenize each complete official
zero-shot prompt, including its context, question, answer choices, and response
format, with the Qwen3.8-27B tokenizer.  We retain the 75 examples whose prompts
contain 90K--128K tokens: 42 labeled hard and 33 labeled easy.  We keep all 42
hard examples and deterministically select eight easy controls.  The easy
examples first add domains and subdomains missing from the hard pool; remaining
slots improve coverage across answer positions and four prompt-length bins.
The final 50 examples cover all six eligible domains and all 13 eligible
subdomains, with answer-position counts of 12, 12, 13, and 13 for A--D.  Prompt
lengths range from 90,049 to 127,548 tokens (108,808 on average).  This
hard-heavy, length-controlled subset is designed for comparison at our context
scale and is not intended to reproduce the distribution of the full benchmark.

\section{API Rates}
\label{app:api-rates}

Table~\ref{tab:api-rates} lists the rates used to convert recorded token usage
into the per-instance costs reported in this paper. Rates are applied to each
API call before costs are averaged across instances.

\begin{table}[h]
  \caption{API rates in USD per million tokens. A dash indicates that we do not
  apply a separate rate for that token category.}
  \label{tab:api-rates}
  \centering
  \small
  \setlength{\tabcolsep}{3pt}
  \begin{tabular}{@{}lrrrr@{}}
    \toprule
    Model & Input & Cache read & Cache write & Output \\
    \midrule
    GPT-5.6-sol             & 4.00 & 0.400 & 5.00 & 20.00 \\
    GPT-5.6-sol (long)      & 8.00 & 0.800 & 10.00 & 30.00 \\
    GLM-5.3                 & 4.86 & 0.972 & 4.86 & 12.15 \\
    Qwen3.8-27B             & 2.48 & -- & -- & 7.46 \\
    Gemini 3.8 Flash        & 0.75 & 0.075 & -- & 3.75 \\
    Kimi-K2.6               & 5.15 & 1.03 & 5.15 & 12.81 \\
    \bottomrule
  \end{tabular}
\end{table}

The GPT-5.6-sol long-context rates apply when a single request exceeds 272K
input tokens. GPT rates follow the
\href{https://developers.openai.com/api/docs/models/gpt-5.6-sol}{OpenAI pricing page},
and Gemini rates follow the
\href{https://cloud.google.com/gemini-enterprise-agent-platform/generative-ai/pricing}{Google pricing page}.
GLM-5.3, Qwen3.8-27B, and Kimi-K2.6 use the rates listed by
\href{https://tinker-docs.thinkingmachines.ai/tinker/models/}{Tinker}.
For the self-hosted Qwen runs, we report a rate-equivalent API cost; all input tokens use the input rate.

\end{document}